\documentclass[letterpaper]{article}
\usepackage{aaai}
\usepackage{times}
\usepackage{helvet}
\usepackage{xcolor}  
\usepackage{courier}
\usepackage{caption}
\usepackage{booktabs}  
\usepackage{tikz}
\usepackage{multirow}
\usepackage{graphicx}
\usetikzlibrary{positioning}
\newcommand{\real}{\mathbb{R}}
\newcommand{\data}{\mathcal{D}}

\newcommand{\ex}{\mathbb{E}}
\newcommand{\ind}{\mathbbm{1}}

\newcommand{\bX}{\mathbf{X}}
\newcommand{\bZ}{\mathbf{Z}}
\newcommand{\bx}{\mathbf{x}}
\usepackage{wrapfig}
\newcommand{\bz}{\mathbf{z}}

\newcommand{\bmu}{\boldsymbol{\mu}}
\newcommand{\bsigma}{\boldsymbol{\sigma}}

\newcommand{\bbeta}{\boldsymbol{\beta}}
\newcommand{\loss}{\mathcal{L}}

\usepackage{algorithmic, algorithm, setspace}
\newcommand{\norm}{\mathcal{N}}

\usepackage{hyperref}
\usepackage{mathtools}
\usepackage{amsmath}
\usepackage{amssymb}
\usepackage{bbm}
\DeclareMathOperator*{\argmax}{arg\,max}

\newcommand\model{\stackrel{\mathclap{\normalfont\mbox{m}}}{=}}
\newcommand\iid{\stackrel{\mathclap{\normalfont\mbox{iid}}}{\sim}}

\begin{document}
%
\title{Diffusing Gaussian Mixtures for Generating Categorical Data}
\author{Florence Regol \and Mark Coates\\ Dept. Electrical and Computer Engineering, McGill University \\Montr\'eal, QC, Canada \\ florence.robert-regol@mail.mcgill.ca, mark.coates@mcgill.ca}
\maketitle
\begin{abstract}
Learning a categorical distribution comes with its own set of challenges. A successful approach taken by state-of-the-art works is to cast the problem in a continuous domain to take advantage of the impressive performance of the generative models for continuous data. Amongst them are the recently emerging diffusion probabilistic models, which have the observed advantage of generating high-quality samples. Recent advances for categorical generative models have focused on log likelihood improvements. In this work, we propose a generative model for categorical data based on diffusion models with a focus on high-quality sample generation, and propose sampled-based evaluation methods. The efficacy of our method stems from performing diffusion in the continuous domain while having its parameterization informed by the structure of the categorical nature of the target distribution. Our method of evaluation highlights the capabilities and limitations of different generative models for generating categorical data, and includes experiments on synthetic and real-world protein datasets.
\end{abstract}

\section{Introduction}

\begin{figure}[bt]
\centering
\includegraphics[scale=0.45]{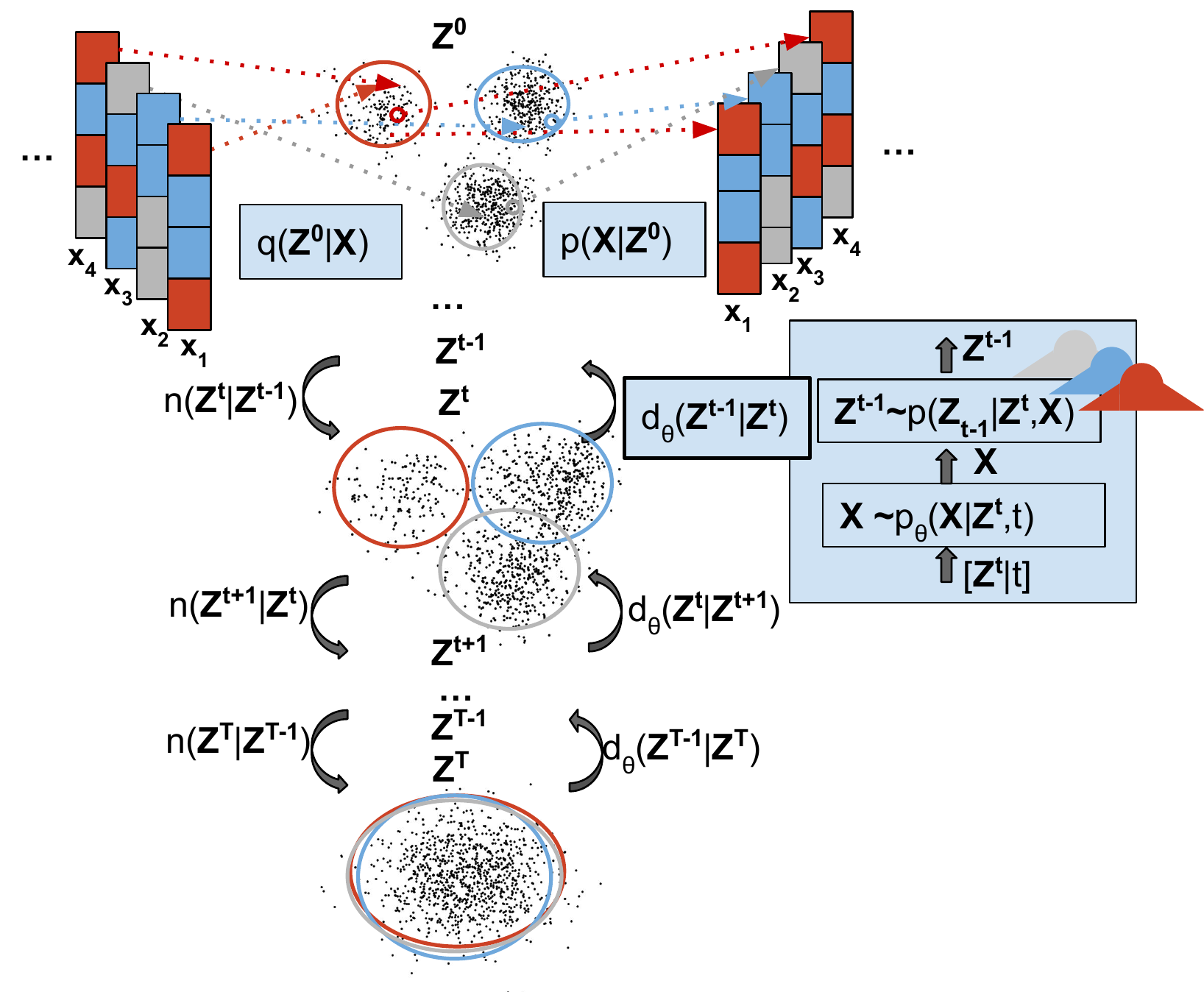}

\caption{Overview of the architecture. On the left, a visualisation of the diffusion process for the first element of a sequence $\bx_{(1)}$ is depicted. The sequence is mapped to the continuous space through the fixed Gaussian Mixture (GM) encoder $q(\bZ^0|\bX=\bx)$, then is diffused through the iterative application of noise distributions $n(\bZ^t|\bZ^{t-1})$ until the signal is destroyed at $\bZ^T$. The right depicts the generative process. Starting from $\bZ^T$, the denoising function models a distribution $\bx$ conditioned on $\bz^T,t$ which in turns models the mixture component of the fixed GM that will be used to produce $\bZ^{t-1}$ conditioned on $\bZ^t, t$. The final sequence is generated from the decoder $p(\bX|\bZ^0)$.}\label{fig:overview}
\end{figure}

There are numerous applications for generative models of categorical random sequences; text generation, speech and music synthesis, drug design and protein synthesis are all important tasks that require modeling of high-dimensional nominal data. 
Learning the structure and substructure underlying those complex high-dimensional distributions can be useful for downstream tasks.
For example, in drug synthesis,  studies have confirmed the important role that mutational covariation plays in determining protein function, and this has found practical applications in drug design and drug resistance prediction~\cite{mcGee2021,tubiana2019,socolich2005}. As a result, recent works have employed generative models to learn from existing proteins and generate new ones~\cite{trinquier2021,mcGee2021,jain2022}. For this type of problem, the ability to generate quality samples is essential.

While the research on generative models for continuous data has been flourishing (see~\cite{bondtaylor2021} for a review), the literature on modeling nominal categorical data is not as developed~\cite{hoogeboom2021}.

Autoregressive (AR) methods  are well suited for modeling categorical data~\cite{cooijmans2017}. 
A notable class of AR models that give impressive performance for this problem are Transformers~\cite{dai2019,child2019,hua2022,jun2020}. Transformers are powerful, but generally suffer from the weaknesses associated with autoregressive models; they are  generally slow to train and slow to sample from~\cite{bondtaylor2021}. They also suffer from quadratic complexity (w.r.t. sequence length), and because of their impressive flexibility in modeling capability, are harder to apply to smaller datasets~\cite{lin2021}.  As a result, many works have attempted to linearize the time/memory complexity~\cite{hua2022,katharopoulos2020,kitaev2020}, but these limitations still remain key challenges~\cite{lin2021}.

Discretization of continuous methods has been explored~\cite{dinh2017,ho2019,theis2016a,uria2013}. However the modeling assumptions of these works are not suited to data that has no natural ordering of the categories. 
For the specific problem of nominal data generation, current state-of-the-art works are based on extending  generative models that were initially developed for continuous data:
normalizing flow~\cite{ziegler2019,lippe2021categorical,hoogeboom2021argmax} and diffusion models~\cite{hoogeboom2021}.~\citeauthor{hoogeboom2021} report results indicating that the diffusion models can outperform Transformers.
Diffusion probabilistic models~\cite{sohldickstein2015deep} are attractive for their generative capability. Compared to their competitors, such models have the characteristic of generating high quality samples and are relatively fast to train. The general trade-off is that they achieve lower likelihood and slower sampling~\cite{bondtaylor2021,ho2020}. As a result, substantial effort has been devoted to address these limitations~\cite{nichol2021,kingma2021,xiao2022,salimans2022}.

 In this work, we propose a generative model based on a diffusion process that can remain in the continuous space without sacrificing our knowledge that the data is nominal. To do so, we introduce a novel approach to encode nominal data in the continuous space via a sphere packing algorithm that places each category in the encoding space. We then incorporate the structural knowledge that follows from this construction into the denoising step of the diffusion using a Gaussian mixture conditioned on the current state of the diffusion. The advantages of such a design are threefold: 1) Unlike previous work~\cite{hoogeboom2021,lippe2021categorical,hoogeboom2021argmax}, this fixed encoding allows flexibility of the dimensionality of the representations without added complexity; 2) the structured denoising step requires significantly fewer diffusion steps, which greatly improves sampling time (which is identified as one of the main limitations of the diffusion model) while keeping the benefit of the diffusion model; 3) the generated samples are of higher quality.

Currently, the main method of evaluating a categorical generative model is via the log likelihood of held-out data. Although useful, this metric has some  known drawbacks. \cite{theis2016a} use a simple example to show clearly how a good likelihood does not guarantee good sample generation. Proper evaluation of generative models is an ongoing research topic in many fields, including text, image, and graph generation~\cite{garbacea2019,celikyilmaz2020,zhou2019,borji2019,thompson2022on,theis2016a,wu2017}. 

The general consensus has been to push towards a more comprehensive and task-oriented approach for assessing performance. Candidate metrics do not necessarily correlate with each other~\cite{theis2016a,zhou2019}, so it can be important to measure performance in multiple ways.
Indeed, failure to follow a comprehensive evaluation methodology has been linked to difficulties in assessing which models are actually better and to unexpected results~\cite{caccia2020,lucic2018,rabanser2019}. A notable example is the finding by~\cite{nagarajan2021} that high likelihood on a dataset and good sample generation does not guarantee good out-of-distribution detection capability, one of the candidate uses of a good generative model. 

With these observations in mind, in this work, we expand on standard evaluation metrics to include distribution distance metrics. 
We propose a synthetic experiment with a known ground truth distribution to aid performance evaluation, with the goal of providing a more complete account of the generative capability of the models considered.
To summarize, the major contributions of this paper are:

\begin{enumerate}
    \item We introduce a \textbf{novel procedure to represent nominal data in the continuous space} based on sphere packing. 
    \item This allows us to design a \textbf{novel denoising function tailored} to model nominal data in the continuous space.
    \item Our presented model offers state-of-the-art \textbf{sample generation quality} and is efficient in both \textbf{sampling time} and \textbf{training time}, as demonstrated by our experiments on both synthetic datasets and on protein datasets.
\end{enumerate}

\section{Related Work}

Early approaches to handle the related problem of discrete data generation were based on dequantization and thresholding. The overall idea is to add noise to the discrete point and treat it as a continuous generative modeling problem, and then use thresholding to generate samples~\cite{ho2019,theis2016a,dinh2017}. 
Current state-of-the-art methods avoid injecting an arbitrary ordering to the categories by either adapting the methodology to stay in the categorical domain, or modelling the data using a latent representation in the continuous space that can be later mapped to the categorical space. In~\cite{ziegler2019} and~\cite{lippe2021categorical}, normalizing flows (NF) are used to model such a latent representation. An encoder-decoder framework is used to map from the categorical to the continuous space and vice versa. The overall model is learned through variational inference.
\cite{hoogeboom2021argmax} build on the same idea as~\cite{lippe2021categorical}, but rather than learning the encoder/decoder, they fix the decoder with an argmax function. This induces a constraint on the functional space of the encoder that is maintained throughout training. Both of these state-of-the-art works keep the mapping from the continuous to the categorical space simple. In~\cite{hoogeboom2021} this is done by using a fixed deterministic argmax function, and~\cite{lippe2021categorical} experimented with learning the encoder/decoder of varying complexity and found that a simple parameterization of the mean and variance gave the optimal result. Unlike our approach, once this mapping is done, nothing informs the NF that it is treating a latent representation of a categorical variable.

Moving away from the normalizing flow methods, ~\cite{hoogeboom2021} also presented a diffusion-based model that operates directly on the categorical space. Instead of diffusing the signal with Gaussian distributions and learning means and variance of parameterized Gaussian as denoising process, they diffuse a one-hot encoded sequence with a multinomial categorical distribution. As is the case for the argmax, the dimension of the sequence representation scales linearly with the number of categories.
Other related work that takes a similar approach to us by mapping to an alternative space to perform diffusion includes~\cite{vahdat2021} and~\cite{sinha2021}.  These works tackle the tangential problems of generating ordinal data and conditional generative modeling.

Lastly, related works that target a similar task connected to generating quality proteins include ~\cite{jain2022,brookes2019,kumar2020,hoffman2022}. This literature focuses on generating high score protein sequences, which are evaluated by an oracle. Even though these models are generative in nature, the end task is still somewhat supervised. The models explicitly aim to maximize a quantity, whereas for our purposes we remain in the traditional generative modeling problem formulation of learning a distribution.

\section{Methodology}
\textbf{Problem Setting.}
Consider a categorical multivariate random variable $\bX = [X_{(1)}, \dots X_{(S)}]$ where each element belongs to one of $K$ categories:  $X_{(j)} \in  \mathcal{C}, \mathcal{C} = \{C_1, \dots,C_K\}$ with associated pmf $p(\bX)$. Given a dataset of realizations $\data = \{\mathbf{x}_i\}^N_{i=1}, \bx_i \sim \bX$, the task is to learn  $p(\bX)$.

\paragraph{Encoding the categorical sequences and sphere packing.} We lift the problem to the continuous space by introducing a latent
continuous random variable $\bZ^0$ that is mapped from and to the
categorical sequence $\bX \in \mathcal{C}^S$ with an encoder,
$q(\bZ^0|\bX)$, and decoder, $p(\bX|\bZ^0)$, respectively. The log
likelihood and its variational lower bound are given by:
{\small \begin{align}
\log p(\bX) &= \log \int \frac{p(\bX, \bZ^0)}{q(\bZ^0|\bX)}q(\bZ^0|\bX)\, d\bZ^0 \nonumber , \\
     \log p(\bX)   &\geq \ex_{q(\bZ^0|\bX)}\Big[ \log \Big(p(\bZ^0) \Big) {+} \log \Big( \frac{p(\bX|\bZ^0)}{q(\bZ^0|\bX)} \Big) \Big] \label{eq:vb1}.
\end{align}}

It is desirable to focus complexity into learning
$p(\bZ^0)$, so we make the mappings from $\bZ^0$ to $\bX$ simple and
tractable.  Consequently, we use a fixed, factorized encoding
distribution to associate each categorical element $X_{(s)}$ of the
sequence with a random vector in a $d$-dimensional continuous space
$\bZ^0_{(s)} \in \real^d $. The mapping depends on the category; each
category $C_k$ is assigned a distribution
$f(\cdot ; \bmu_{C_k}, \sigma)$ that is clearly distinguishable
from others by its mean $\bmu_{C_k} \in \real^d$ and variance
$\sigma^2 \in \real $. We use a Gaussian $f(\cdot)$ for
simplicity, and similarly to~\cite{lippe2021categorical} we obtain the decoder $p(\bX|\bZ^0)$ through Bayes'
rule, so we have:
{\small \begin{align}
    q(\bZ^0|\bX) &= \prod^S_{s=1}   \norm(\bZ^0_{(s)}  ; \bmu_{X_{(s)}}, \mathbf{I}\sigma^2 ) \text{ as the encoder and} \nonumber  \\
    p(\bX|\bZ^0) &= \prod^S_{s=1}  \frac{  \norm(\bZ^0_{(s)}  ; \bmu_{X_{(s)}}, \mathbf{I}\sigma^2) }{\sum^K_{k =1}  \norm(\bZ^0_{(s)}  ; \bmu_{C_k}, \mathbf{I}\sigma^2) }\text{ as the decoder}. \nonumber  
\end{align}}
(The prior on $\bX$ does not appear as we assume uniformity).
The advantages are twofold: 1)  it imposes a
structure on the target distribution $p(\bZ^0)$ that can be used in
modeling the learnable $p_{\theta}(\bZ^0)$, as we will show shortly; and 2) it simplifies the learning objective since only $p(\bZ^0)$ is learnable.

Our aim is to make it as easy as possible for the decoder to
distinguish between categories. This implies that we should strive
to identify maximally separated means. This leads to a sphere
packing problem --- finding the emplacement of $K$ points on the
surface of a $d$-dimensional sphere $\mathbb{S}^d(1) $ that
maximizes the minimum distance between any two points:
{\small \begin{align}
   \bmu^*_1 ,\dots, \bmu^*_K &=  \argmax_{\bmu_1 ,\dots, \bmu_K \in \mathbb{S}^d(1)} \quad \Big(\min_{i \neq j} ||\bmu_i - \bmu_j||^2_2 \Big) \nonumber
\end{align}}
Hence we can use solutions of this problem, e.g.,~\cite{gamal1987}, to 1) set the means
of the encoding distributions
$\{\bmu_{C_k}\}^K_{k=1}$; and 2)
determine, based on the minimum distance $d_{\bmu^*}= \min_{i \neq j} ||\bmu^*_i - \bmu^*_j||^2_2 $, a value for the
variance $\sigma^2$ such that the Gaussian distributions
$\norm(\bmu_{C_k},\mathbf{I}\sigma^2); k \in [K]$ have
limited overlap but are not too concentrated. Denoting $d_{\bmu^*} =
\min_{i \neq j} ||\bmu^*_i - \bmu^*_j||^2_2$,  we have:
{\small \begin{align}
   \bmu_{C_k} = \bmu^*_{k} ; \quad k \in [K] \quad 
                 \text{and} 
                 \quad 
   \sigma = \frac{d_{\bmu^*}}{2K\sqrt[d]{3}}.
\end{align}}
Almost all ($99.7\%$) of the mass of a $d$-dimensional m.v. Gaussian R.V. is within $\sqrt[d]{3}$ standard deviations,
so we set $\sigma$ to half that radius, and divide by the number of categories.

\textbf{Learning the latent distributions $p_{\theta}(\bZ^0)$.} The complex correlation structure of the categorical distribution must be captured in $p_{\theta}(\bZ^0)$. We propose to use a diffusion probabilistic model (DPM)~\cite{sohldickstein2015deep} with a novel denoising component, tailored to our encoding scheme and categorical data, based on Gaussian Mixtures.
The DPM introduces $T$ latent random variables $\bZ^1, \dots, \bZ^T$. Commencing with the targeted encoded sequence $\bZ^0$, the variables are derived by gradually adding known Gaussian noise of increasing variance to the variable from the previous timestep:
$n(\bZ^t|\bZ^{t-1}) = \norm(\bZ^t ; \sqrt{1-\beta_t}\bZ^{t-1},\beta_t \mathbf{I}); \beta_i < \beta_{i+1} \in (0,1)$. At the end of the chain, only noise should remain $\bZ^T \sim \norm(\bZ^T; \textbf{0},\textbf{1})$. The task of the DPM is to learn the denoising process $d_{\theta}(\bZ^{t-1} | \bZ^t); t \in [T]$. 

This leads to construction of the generative model for $\bZ^0$:
{\small \begin{align}
    p_{\theta}(\bZ^0) \model d_{\theta}(\bZ^0) &=  \int d(\bZ^T) \prod^T_{t=1} d_{\theta}(\bZ^{t-1} | \bZ^t)d\bZ^{1:T}. \label{eqn:diff_trans_likelihood}
   \end{align}}
 See~\cite{sohldickstein2015deep,ho2020} for more detailed discussion of the diffusion process. 

 \paragraph{Exploiting the structure.} In
 most denoising approaches, the distributions $d_{\theta}(\bZ^{t-1} |
 \bZ^t)$ are modelled as normal distributions with learnable means and (usually fixed) variances. In our case, we take advantage of the known structure. By our
 construction, the target distribution is a mixture of
 Gaussians; conditioned on knowledge of the target sequence, the
 distribution $p(\bZ^{t-1} | \bZ^t,\bX)$ is Gaussian, and the
 mean and variance can be evaluated analytically.
 
If we are at a point in the chain $\bz^t$, then if we are given an element of
 the sequence $x_{(s)}$, $\bZ^{t-1}_{(s)}$ is conditionally independent of other $\bZ^{t-1}_{(s')}$, and we can derive the conditional of the next
 denoising step in closed-form: 
{\small\begin{align}
    p(\bZ^{t-1}_{(s)} | \bZ^{t}_{(s)} , x_{(s)}) &=  \int p(\bZ^{t-1}_{(s)} | \bZ^{t}_{(s)} , \bZ^{0}_{(s)}) p( \bZ^{0}_{(s)}| x_{(s)}  )  d\bZ_{(s)}^0 \nonumber
     \\
     &= \norm(\bZ^{t-1}_{(s)}; \bmu^{\bZ^t,t}_{x_{(s)}}, \mathbf{I}\sigma^2_{t}) \label{eq:posterior} \\
    \text{ where } \bmu^{\bZ^t,t}_{x_{(s)}} = &\frac{\sqrt{\bar{\alpha}_{t-1}}\beta_t}{1-\bar{\alpha}_t}\bmu_{x_{(s)}} +  \frac{\sqrt{\alpha_t}(1-\bar{\alpha}_{t-1})}{1-\bar{\alpha}_t}\bZ_{(s)}^t, \nonumber \\
    \sigma^2_{t} =&    \frac{1-\bar{\alpha}_{t-1}}{1-\bar{\alpha}_t}  \beta_t+ (\frac{\sqrt{\bar{\alpha}_{t-1}}\beta_t}{1-\bar{\alpha}_t} \sigma)^2  \nonumber.
\end{align}}
(See the supplementary for the detailed derivation.) Hence if we have a predictor $p_{\theta}(\bX|\bZ^t, t)$ of the
distribution of the sequence $\bX$ based on the current state $\bz^t$
and the diffusion step $t$, we can model the denoising step as:
{\small \begin{align}
d_{\theta}(\bZ^{t-1} | \bZ^t) =  \sum_{\bX \in \mathcal{C}^S} \left(\prod_{s=1}^S p(\bZ^{t-1}_{(s)} | \bZ^{t}_{(s)} , X_{(s)})
                                   \right)p_{\theta}(\bX|\bZ^t, t)\,.
\end{align}}
If $p_{\theta}(\bX|\bZ^t, t)$ is structured to assume independence
among the elements of $\bX$, we can factorize $p_{\theta}(\bX|\bZ^t, t) =
\prod_{s=1}^S p_{\theta}(X_{(s)}|\bZ^t, t)$ and write:
\begin{align}
   d_{\theta}(\bZ^{t-1} | \bZ^t) &=
  \prod_{s=1}^S \sum^K_{k=1} p(\bZ^{t-1}_{(s)} | \bZ^{t}_{(s)} ,C_k)  p_{\theta}(X_{(s)}=C_k|\bZ^t, t)\, \nonumber
  \\&= \prod_{s=1}^S d_{(s),\theta}(\bZ_{(s)}^{t-1} | \bZ^t). \label{eqn:rev_trans_simp} 
\end{align} 
Replacing the Gaussian denoising term used in~\cite{sohldickstein2015deep,ho2020} with this more complex denoising model results in a more involved loss expression, but the denoising process can be successful with far fewer diffusion steps (10-40 versus thousands). This effect was also observed in~\cite{xiao2022}.

\paragraph{Loss objective.}

Since the encoder and decoder are fixed, optimization of the loss
function (Eqn.~\eqref{eq:vb1}) simplifies to:
{\small\begin{align}
\theta^* =\argmax_{\theta \in \Theta} \ex_{q(\bZ^0|\bX)}\Big[ \log \Big(p_{\theta}(\bZ^0) \Big)  \Big]\,,  \label{eq:general_loss}
\end{align}}
i.e., the log likelihood of the diffusion model under the expectation of the encoder. Since the DPM is a latent variable model, its log likelihood is also optimized via a lower bound:
{\small\begin{align}
    \log &\Big(p_{\theta}(\bZ^0) \Big) 
    \geq =\mathbb{E}_{n}\Big[ - \log \frac{ n\left(\mathbf{Z}^{T} \mid \mathbf{Z}^{0}\right)}{d\left(\mathbf{Z}^{T}\right)} \nonumber\\ &-\sum_{t=2}^T  \log \frac{n\left(\mathbf{Z}^{t-1} \mid \mathbf{Z}^{t}, \mathbf{Z}^{0}\right)}{   d_{\theta}(\bZ^{t-1} | \bZ^t)} +\log d_{\theta}\left(\mathbf{Z}^{0} \mid \mathbf{Z}^{1}\right)\Big] .\nonumber
\end{align}}

Employing this bound in Eqn.~\eqref{eq:general_loss}, substituting
with~\eqref{eqn:rev_trans_simp}, and removing
terms that are independent of $\theta$, we can identify the final
optimization task:
{\small\begin{align}
\theta^* &= \argmax_{\theta \in \Theta} \ex_{q,n}\Big[ -\sum_{t=2}^T   \loss_{t-1} +  \loss_0 \Big]\, \label{eq:final_loss} 
\end{align}}
where $\loss_{t-1} = KL\Big(n\left(\mathbf{Z}^{t-1} \mid \mathbf{Z}^{t}, \mathbf{Z}^{0}\right)||d_{\theta}(\bZ^{t-1} | \bZ^t)\Big)$
and $\loss_0 = \log d_{\theta}\left(\mathbf{Z}^{0} \mid \mathbf{Z}^{1}\right) $.

\paragraph{Architecture and training.} 
In practice, it has been shown beneficial for this type of loss to randomly optimize one of the terms $\loss_t$ at a time~\cite{ho2020} ~\cite{nichol2021}.  The objective then becomes to either maximize the log likelihood of the final step for $t=0$, or to minimize the KL divergence between a Gaussian mixture with learnable mixture weights for time step $t>0 $:
{\small \begin{align}
 \loss_{t-1}  &=KL\Big(n\left(\mathbf{Z}^{t-1} \mid \mathbf{Z}^{t}, \mathbf{Z}^{0}\right)||d_{\theta}(\bZ^{t-1} | \bZ^t)\Big) \nonumber \\
 &=  \sum^S_{s=1} KL\Big(n\left(\mathbf{Z}_{(s)}^{t-1} \mid \mathbf{Z}_{(s)}^{t}, \mathbf{Z}_{(s)}^{0}\right)|| d_{(s),\theta}(\bZ_{(s)}^{t-1} | \bZ^t) \Big)  \nonumber 
 \end{align} }
 Using the variational approximation of the KL divergence between Gaussian mixtures from~\cite{hershey2007}, we can approximate the individual step loss as follow:
 {\small
 \begin{align}
       \loss_{t-1}  &\approx - \sum^S_{s=1}  \log  \sum^K_{k=1 } p_{\theta}(X_{(s)}=C_k|\bZ^t, t) w_{\bZ^{t,0}}^s(C_k) \nonumber \\
     \text{where }& w_{\bZ^{t,0}}^s(C_k) = \exp^{-KL\left(  n(\cdot  \mid \mathbf{Z}_{(s)}^{t}, \mathbf{Z}_{(s)}^{0}) || \norm(\cdot; \bmu^{\bZ^t,t}_{C_k}, \sigma^2_{t} \mathbf{I}) \right) }.  \nonumber
\end{align} }
Details of the derivation are provided in the supplementary. 

At this point, we can see that the optimization of this term is reached when $p_{\theta}(X_{(s)}|\bZ^t, t)$ gives maximum weight to the highest term of the sum $ w_{\bZ^{t,0}}^s(C_k)$, which is the initial sequence $C_k = x_{(s)}$. As a result, we approximate this optimization by maximizing the log likelihood of  $p_{\theta}(X_{(s)}=x_{(s)}|\bZ^t, t)$ , as both isolated optimization problems have the same solution:
 \begin{align}
       \argmax_{\theta \in \Theta}  \loss_{t-1}    & \approx  \argmax_{\theta \in \Theta}   \log  p_{\theta}(\bX =  \bx |\bZ^t, t) \label{eqn:loss_almost_practice}  
 \end{align}
As a result, learning hinges on the modeling capability of $p_{\theta}(\bX|\bZ^t, t)$. We employ a
transformer-based architecture. The vector
$\bz^t$ and an embedding of time $t$ serve as inputs.  We adopt a
sampling approach for the training. For each
sequence in the training data, we sample
$\bz^0 \sim q(\bZ^0|\bX = \bx)$, and then draw a time $t$, we sample $\bz^t \sim n(\bZ^t|\bz^0)$ to evaluate the loss.
It is important to emphasize that this transformer does {\em not} have an autoregressive structure --- all elements of a sequence are generated in parallel. The correlations are induced by the denoising diffusion process.

\begin{figure}[bt]
    \centering
\begin{minipage}{\textwidth}
\begin{tikzpicture}
  \node (img)  { \includegraphics[scale=0.35]{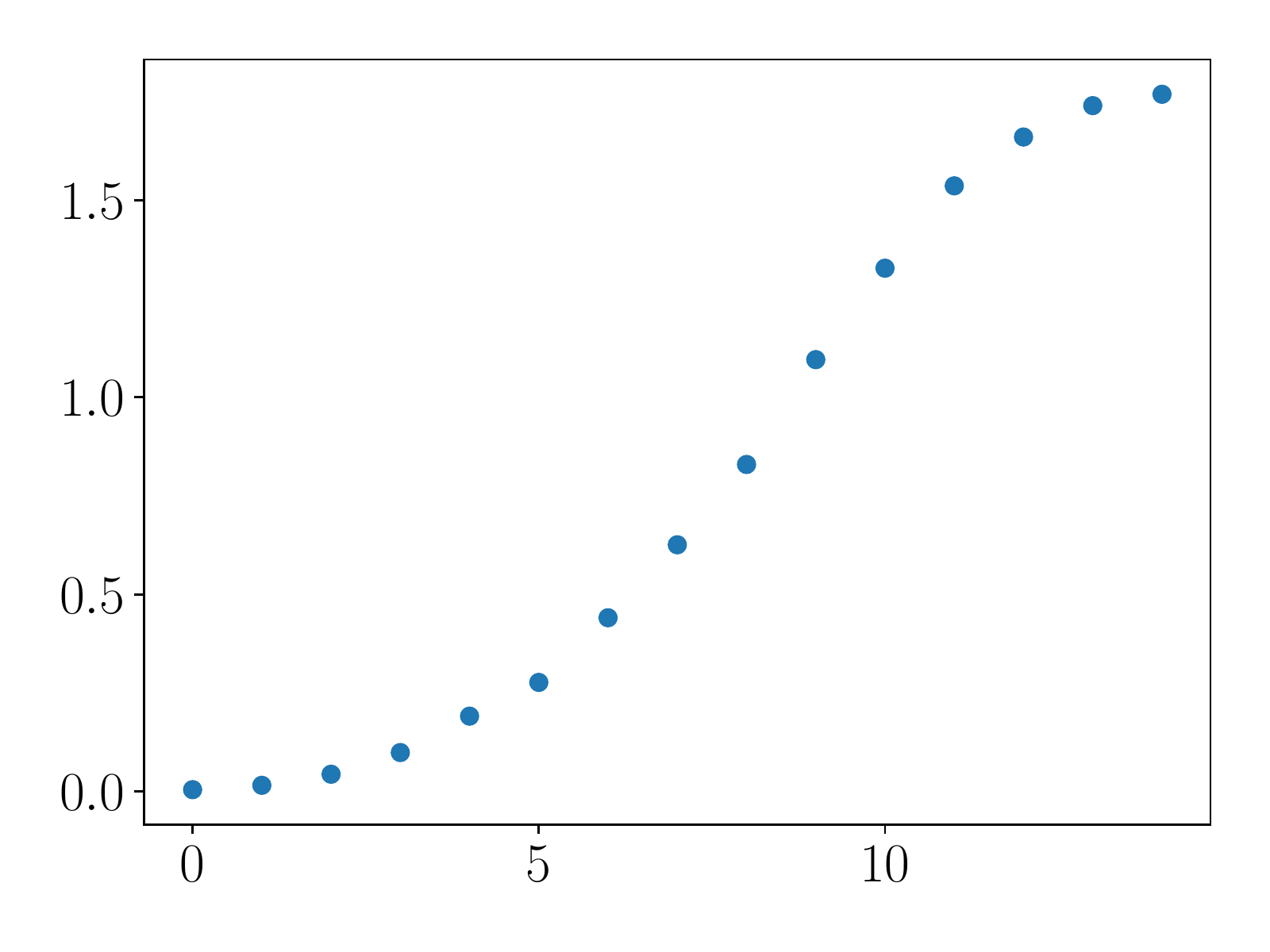}
    };
  \node[below=of img, node distance=0cm, yshift=1.5cm] { $t$};
  \node[left=of img, node distance=0cm, rotate=90, anchor=center, yshift=-1cm] {$H(p_{\theta}(\bX | \bZ^t, t))$};
\end{tikzpicture}
\end{minipage}

\caption{Average entropy of $p_{\theta}(\bX | \bZ^t, t)$ along the diffusion process during sampling. The model becomes more and more certain (lower entropy) as we approach $t=0$.}  \label{fig:entropy_output}
   
\end{figure}

\paragraph{Data Augmentation.} 
In practice, we observe that $p_{\theta}(\bX | \bZ^t, t)$ learns to be increasingly certain of its prediction as we approach the end of the chain $\bZ^0$. This behavior can be seen in Figure~\ref{fig:entropy_output} where we show an example of the entropy at every time step $H(p_{\theta}(\bX | \bZ^T, T)), \dots, H(p_{\theta}(\bX | \bZ^1, 1))$.

We can imagine that alongside the gradually noisy $\bZ^t$, there is also a corresponding noisy categorical sequence $\tilde{\bX}^t$ that $p_{\theta}(\bX | \bZ^t, t)$ aims to predict.
As a result, instead of training on the ground truth sequence at the beginning of the diffusion $T, T-1 ,\dots$, we inject some noise by training on a ``diffused'' version of $\bx$, denoted by $\tilde{\bx}^t$, and thus modify Eqn.~\eqref{eqn:loss_almost_practice}  to:
\begin{align}
       \argmax_{\theta \in \Theta}  \loss_{t-1}    & \approx  \argmax_{\theta \in \Theta}   \log  p_{\theta}(\bX =  \tilde{\bx}^t |\bZ^t, t) \label{eqn:loss_practice} \\
         \tilde{\bx}^t \sim p^{\bz^t, t }(\tilde{\bX})& \text{, where }
    p^{\bz^t, t }\big(\tilde{\bX} \big)_s = \frac{ (w_{\bZ^{t,0}}^s(x_{(s)}) )^{\omega}}{\sum^K_{k=1} (w_{\bZ^{t,0}}^s(C_k) )^{\omega}}  \nonumber 
\end{align}
Algorithms detailing the training and sampling procedures are provided in the supplementary.

\section{Experiments}
In this section, we first present the evaluation metrics, the datasets and the experimental set-up. We then report the performance of our proposed GMCD model and conduct ablation studies to validate the effectiveness of its modules.

\paragraph{Evaluation metrics.}

{\small\begin{table*}[bt]  \footnotesize \centering
\begin{tabular}{llcccc|cc|c}  \toprule
&&$Hel$ &$d_{TV}  $ &($d_{TV+}$ &$ d_{TVood})$ &$p(A_{likely})$ &$p(A_{rare})$ &  $p(A_+)$ \\ \midrule
\multirow{4}{*}{ \rotatebox[origin=c]{90}{$\mathbf{K=6}$}}&CNF+ &$37.02 $ &$34.63 $ &$25.59 $ &$9.04 $ &$40.91 $ &$41.01 $ &$81.92 $ \\ 
&argmaxAR+ &$24.22 $ &$17.90 $ &$13.55$ &$4.35 $ &$66.32 $ &$24.97$ &$91.29 $ \\ 
&CDM &$19.27 $ &$16.19 $ &$14.34 $ &$1.84 $ &$66.33 $ &$29.98 $ &$96.31 $ \\ 
&GMCD &$\mathbf{16.62}*$ &$\mathbf{16.07}$ &$15.56 $ &$0.51$ &$\mathbf{75.71}*$ &$\mathbf{23.27}* $ &$\mathbf{98.98}*$ \\ 
\toprule
\multirow{4}{*}{ \rotatebox[origin=c]{90}{$\mathbf{K=8}$}}&CNF+ &$81.86 $ &$83.63 $ &$69.79 $ &$13.84 $ &$35.98 $ &$36.34 $ &$72.32 $  \\
&argmaxAR+ &$73.48 $ &$76.06 $ &$73.19$ &$2.87 $ &$66.94 $ &$27.32 $ &$94.27 $ \\
&CDM &$73.65 $ &$76.36 $ &$73.81 $ &$2.55 $ &$62.70 $ &$32.20 $ &$94.90 $ \\ 
&GMCD &$\mathbf{72.30}*$ &$\mathbf{74.98}*$ &$73.34 $ &$1.64$ &$\mathbf{71.77}*$ &$\mathbf{24.94}*$ &$\mathbf{96.71}*$ \\ 
\toprule
\multirow{4}{*}{ \rotatebox[origin=c]{90}{$\mathbf{K=10}$}}&CNF+ &$98.28 $ &$99.81 $ &$83.43 $ &$16.38 $ &$33.64 $ &$33.59 $ &$67.23 $\\
&argmaxAR+ &$98.43$ &$99.81 $ &$76.95 $ &$22.86 $ &$40.38 $ &$13.89 $ &$54.27 $ \\
&CDM &$97.32 $ &$99.69 $ &$97.39$ &$2.30 $ &$64.62 $ &$30.77 $ &$95.40 $ \\ 
&GMCD &$\mathbf{97.27}*$ &$\mathbf{99.68}*$ &$97.71 $ &$1.98$ &$\mathbf{66.99}*$ &$\mathbf{29.05}*$ &$\mathbf{96.05}*$ \\
\toprule
&optimal  & $0$ &$0$ & & &$75$ &$25$ &$100$ \\ 
\bottomrule
\end{tabular} \caption{Distances metrics and probability estimates for partitionings $\mathcal{P},\mathcal{P}^{od}$  for the synthetic datasets.$*$ indicates significance w.r.t. to the Wilcoxon  signed-rank test at the $5\%$ level. $+$ indicates that more epochs were required to reach competitive results.} \label{tab:synth} \end{table*}}

{\small \begin{table}[b]
  \footnotesize   \centering
\begin{tabular}{lcccc} \toprule
& argmaxAR+ &  CNF+      & CDM       &  GMCD      \\ \midrule
num. params          & $250K$                        & $180K$                   & $\mathbf{40K}$                   & $\mathbf{40K}$                    \\
epoch time           & 1.9x                          & 1.6x                     & \textbf{1x}                      & \textbf{1x }                      \\ 
sampling time        & 1.2x                          & 1.2x                     & 1.1x                    & \textbf{1x}                       \\ \bottomrule
\end{tabular}  \caption{Timing with $K=8$. Experiments are conducted on GPU machines NVIDIA GeForce RTX 2060
.}
    \label{tab:timing}
\end{table}}
Many of the difficulties and limitations associated with evaluating generative models stem from the fact that we do not have access to the ground truth distribution.
With access to  ground truth, the problem formulation changes and the previously mentioned problems associated with log likelihood (LL) and sampled-based metrics disappear. Instead of:
\begin{itemize}
\item Maximizing the LL of unseen samples $\rightarrow$, we aim to assign the correct probability mass to unseen samples,
\item Generating ``good'' samples $\rightarrow$ we aim to generate samples that are distributed according to the ground truth,
\item Maximizing a heuristic for sample quality (novelty, diversity, etc.) $\rightarrow$ we aim to generate samples with the same heuristic value as the expected value from ground truth. 
\end{itemize}
In this work, we are interested in evaluating how close a generative model is to the true probability measure based on its samples in the discrete domain.

\paragraph{With known ground truth distribution.}
The distance between two distributions $p,q$ on a discrete sample space $\Omega$ can be measured by the total variation and Hellinger distances:
{\small \begin{align}
    d_{TV}(p,q) & \triangleq \frac{1}{2}||p-q||_1 =  \frac{1}{2} \sum_{x \in \Omega} |p_x -q_x| \nonumber, 
\\Hel(p,q) &\triangleq \frac{1}{\sqrt{2}}||\sqrt{p}-\sqrt{q}||_2 = \frac{1}{\sqrt{2}} \sqrt{\sum_{x \in \Omega} (\sqrt{p_x} - \sqrt{q_x})^2}. \nonumber
\end{align}}
(with $p_x$ used as a shorthand for $p(x)$). These are principled metrics but they can rapidly become impractical as $\Omega$ grows, especially as we must usually rely on samples to estimate $p_x$.
Alternatively, we can consider a partitioning of the sample space: $\mathcal{P} = \{A_i ; A_i \subset \Omega, A_i \cap A_j = \emptyset \} $ and estimate the probability mass of these events $p(A_i) = \sum_{x \in A_i} p_x$. It is less precise but can be more informative if $\Omega$ is large and/or if the partitioning has a particular meaning. 
One obvious partitioning of interest would be to divide the sample space into positive-support elements (in distribution - ID) and the zero support elements (out-of-distribution - OOD); the partitioning is then
$\mathcal{P}^{od} = \{A_o, A_{+} \} $; where $\{x \in A_{+} ; p_x > 0,\,\, x \in \Omega \}$, $\{x \in A_o ; p_x = 0,\,\,x \in \Omega \}$.
 
 As our focus is on sample quality, we compare the ground truth distribution $p$ to the empirical distribution $\hat{p}_{\theta}$ constructed from the samples of a generative model. 
 For the synthetic experiments where we have access to $p$, we report:
 {\small \begin{itemize}
     \item $Hel(p,\hat{p}_{\theta})$ and $d_{TV}(p,\hat{p}_{\theta})$,
     \item $d_{TV+} \triangleq  \frac{1}{2} \displaystyle{\sum_{x \in A_+}} |p_x -\hat{p}_{\theta x}| $ , $d_{TVo}\triangleq  \frac{1}{2} \displaystyle{\sum_{x \in A_o}} |p_x -\hat{p}_{\theta x}| $ ,
     \item  $\hat{p}_{\theta}(A_{+}) = \displaystyle{\sum_{x \in A_+} }\hat{p}_{\theta x} $ ; prob. estimates of valid sequences,
     \item $\hat{p}_{\theta}(A_i)=\displaystyle{\sum_{x \in A_i} }\hat{p}_{\theta x}$ ; prob. estimates of specified $A_i$.
     \end{itemize}} 
\paragraph{Without ground truth distribution.}
In practice, $p$ is not available. We still focus on generating samples that are representative of the distribution by comparing statistics of the ground truth distribution with those derived from generated samples. 
A major capability of interest of a generative model is its ability to properly capture patterns in the data; as such we can compare the higher order covariation of patterns of a generated set of samples to that of a test set. Such evaluation metrics are commonly used in the generative protein sequence modeling literature~\cite{trinquier2021,mcGee2021}. 
Given a pattern of size $p$, described by positions and corresponding categories $ (\{s_1, \dots, s_p\},\{k_1,\dots, k_p\})$, and a set of $M$ sequences $\bx^M$, the higher order pattern covariation $C^{s_1, \dots, s_p}_{k_1,\dots, k_p}(\bx^M)$ is the frequency of the appearance of the pattern in $\bx^M$ minus the product of the frequencies of each individual element of the pattern:
{\small \begin{align}
  \hat{f}^{s_1, \dots, s_p}_{k_1,\dots, k_p}(\bx^M) &= \frac{1}{M}\sum^M_{i=1} \ind[x^i_{(s_1)} = k_1, \dots, x^i_{(s_p)} = k_p]\,, \nonumber \\
                                                      C^{s_1, \dots, s_p}_{k_1,\dots, k_p}(\bx^M) &= \hat{f}^{s_1, \dots, s_p}_{k_1,\dots, k_p}(\bx^M) - \prod^p_{j=1} \hat{f}^{s_j}_{k_j}(\bx^M). \label{eqn:cov} 
\end{align}}
For a given pattern length $p$, we select a random subset of all possible patterns $\{pattern^p_1, \dots \}$ by following the procedure described in~\cite{mcGee2021}, which focuses on the most likely patterns (the detailed selection procedure is described in the supplementary). We report the Pearson correlation $\rho^p$ between the pattern higher order covariations computed on the test set $\mathbf{C}_{p}= [ C^{pattern^p_1}(\bx), \dots ], \bx \sim \data $ and the set of generated samples $\mathbf{C}_{\hat{p}_{\theta}}= [ C^{pattern^p_1}(\bx), \dots ], \bx \sim \hat{p}_{\theta}$.

\subsection{Datasets}
{\begin{table*}[bt] \footnotesize \centering 
\begin{tabular}{llcccccccc} \toprule
&&$\rho^2$ &$\rho^3$ &$\rho^4$ &$\rho^5$ &$\rho^6$ &$\rho^7$ &$\rho^8$ &$\rho^9$ \\ 
\toprule
\multirow{4}{*}{ \rotatebox[origin=c]{90}{\textbf{PF00076}}}&CNF&- &- &- &- &- &- &- &- \\ 
&argmaxAR &$73.13 $ &$73.00 $ &$69.85 $ &$63.88 $ &$58.74 $ &$49.08 $ &$49.74 $ &$56.03 $ \\ 
&CDM &$82.30 $ &$82.44 $ &$80.48 $ &$78.08 $ &$74.95 $ &$73.10 $ &$75.56 $ &$77.27 $ \\ 
&GMCD &$\mathbf{84.19}*$ &$\mathbf{82.85}$ &$\mathbf{82.36}*$ &$\mathbf{81.04}*$ &$\mathbf{77.67}*$ &$\mathbf{78.09}*$ &$\mathbf{78.95}*$ &$\mathbf{80.39}*$ \\ 
\toprule
\multirow{4}{*}{ \rotatebox[origin=c]{90}{\textbf{PF00014}}}&CNF&- &- &- &- &- &- &- &- \\ 
&argmaxAR &$78.06 $ &$79.05 $ &$80.89 $ &$83.57 $ &$84.97 $ &$\mathbf{88.51}$ &$91.26 $ &$91.46 $ \\ 
&CDM &$\mathbf{81.28}*$ &$80.48 $ &$78.98 $ &$78.76 $ &$77.35 $ &$80.40 $ &$85.31 $ &$89.90 $ \\ 
&GMCD &$80.41 $ &$\mathbf{80.81}*$ &$\mathbf{82.01}*$ &$\mathbf{84.09}*$ &$\mathbf{85.83}$ &$88.39 $ &$\mathbf{91.50}$ &$\mathbf{93.04}$ \\ \hline
\multirow{2}{*}{ \rotatebox[origin=c]{90}{\textbf{abl.}}} &GMCD random  &$69.21$ &$69.04$ &$71.69$ &$78.81$ &$80.26$ &$87.65$ &$90.54$ &$93.20$ \\ 
&GMCD sharp &$79.86$ &$79.13$ &$81.07$ &$82.96$ &$83.85$ &$85.26$ &$88.78$ &$89.80$  \\ \bottomrule
\end{tabular} \caption{ Proteins experiment results.  $-$ indicates that the pearson coefficient was not significant at the $5\%$ level.}\label{tab:prot} \end{table*}}

{\small\begin{table}[bt] \footnotesize \centering 
\begin{tabular}{llcccc} \toprule
&&$\rho^2$ &$\rho^3$ &$\rho^4$ &$\rho^5$ \\ \toprule
\multirow{4}{*}{ \rotatebox[origin=c]{90}{$\mathbf{K=6}$}}&argmaxAR+ &$63.16 $ &$58.38 $ &$59.22 $ &$63.66 $ \\ 
&CNF+ &$-12.97 $ &$14.61 $ &$-5.05 $ &$-21.10 $ \\ 
&CDM &$54.00 $ &$58.26 $ &$59.07 $ &$63.52 $ \\ 
&GMCD &$\mathbf{64.03}$ &$\mathbf{63.88}$ &$\mathbf{66.22}$ &$\mathbf{67.64}$ \\ 
\toprule
\multirow{4}{*}{ \rotatebox[origin=c]{90}{$\mathbf{K=8}$}}&argmaxAR+ &$30.93 $ &$21.49 $ &$13.90 $ &$\mathbf{14.24}$ \\ 
&CNF+ &  $-10.13 $ &$-$ &$3.20 $ &$-1.31 $ \\
&CDM &$20.19 $ &$16.61 $ &$12.03 $ &$6.98 $ \\ 
&GMCD &$\mathbf{32.70}$ &$\mathbf{26.83}*$ &$\mathbf{16.31}$ &$10.71 $ \\  \toprule
\multirow{4}{*}{ \rotatebox[origin=c]{90}{$\mathbf{K=10}$}}&argmaxAR+ &$11.85 $ &$6.54 $ &$4.57 $ &$1.55 $\\
&CNF+ & $- $ &$-4.68 $ &$-1.62 $ &$-$ \\
&CDM &$22.46 $ &$13.40 $ &$\mathbf{7.80}$ &$4.55 $ \\ 
&GMCD &$\mathbf{25.19}$ &$\mathbf{18.67}$ &$6.60 $ &$\mathbf{4.79}$ \\ 
\bottomrule 
\end{tabular}\caption{ Pattern covariance metrics.}\label{tab:synth_rho} \end{table}}
We design a ground truth distribution to generate a synthetic dataset of sequences of length $S=K$. We define the sample space $\Omega^K =  \mathcal{C}^K$ and only assign probability mass on permutations of $\mathcal{C}$, i.e.,  $A_{+} = \{\bx; x_{(i)}\neq x_{(j)} \,\, \forall i\neq j\}$. Finally, we separate the positive sets in two and assign  3 times more mass to sequences with a ``smaller'' category at the start of the sequence than at the end,  i.e.:
{\small \begin{align}
    p(\bx ) = \begin{cases} 
    \frac{3}{2 K!}    &  \text{if }\bx \in A_{likely} = \{\bx; \bx \in A_{+} \land x_{(1)}<x_{(S)} \}, \\
\frac{1}{2 K!}  &  \text{if }\bx \in  A_{rare}  = \{\bx; \bx \in A_{+} \land x_{(1)}>x_{(S)} \},   \\
    0 & \text{otherwise}\,. \\ 
    \end{cases}  \nonumber 
\end{align}}
This synthetic dataset is designed to emulate characteristics of a real world dataset. In practice, the distributions that we wish to model are likely to have positive support on a very small fraction of the probability space. Whether we are trying to generate text, images or proteins,  the likelihood of stumbling across a ``valid'' sample when drawing from a uniform distribution is extremely small.

Natural partitionings of interest for this type of dataset are: 1) $\mathcal{P}^{od}$ as previously described where we can see a model's ability to grasp the positive support of the sample space; and 2)
$\mathcal{P} = \{A_{likely}, A_{rare}, A_o\}$ where we can see a model's ability to assign the right amount of probability mass to the different sets.

{\small \begin{table} [b] \footnotesize
    \label{tab:data_synth}
    \resizebox{\columnwidth}{!}{
    \centering \footnotesize
    \begin{tabular}{lllc} \toprule
     $S,K$ & $|\Omega|$ &  $|A_{+}|$ & \% $A_+$ in training set \\ \midrule
        $6$ & $6^6 = 46,656$ &$6!=720$ & $100\% $\\
        $8$ & $8^8 = 16,777,2164$ & $8!=40, 320$ & $21.34\% $\\
       $10$  & $ 10^{10}$ &$3,628,800$&$0.28\%$\\ 
       \bottomrule
    \end{tabular}}
\caption{Size of the sample  space $|\Omega|$, of the positive support set $|A_{+}|$(number of valid sequences) and the fraction of valid sequence contained in the training set of the synthetic datasets. We generate 10K sequences for the train/valid/test set for a total size $N=30K$.}\end{table}}
We consider a small scale experiment $K=6$ where the models are exposed to the entire ID set $A_+$ multiple times, a medium scale experiment $K=8$ where the models are exposed to a sizeable fraction of the ID set, and a larger scale experiment $K=10$ where the models are exposed to less than $1\%$ of $A_+$ (see Table~\ref{tab:data_synth} for additional details).

As a real world application, we measure the performance of the models on two protein datasets from the Pfam protein family~\nocite{ElGebali2019}: \textbf{PF00076}, which contains $N=137,605$ proteins of length $S=70$ and \textbf{PF00014},  which contains $N=13,600$ proteins of length $S=53$. The number of categories for both datasets corresponds to the list of amino acids $K=21$.

\subsection{Experiment Details}

\textbf{Baselines.}We compare our GMCD approach to three state-of-the-art baselines; 1) CNF~\cite{lippe2021categorical}, a normalizing flow method that learns a mapping to/from the categorical space;  2) CDM~\cite{hoogeboom2021}, a diffusion-based model; and 3) argmaxAR~\cite{hoogeboom2021argmax}, a normalizing flow method that uses an argmax operation to map to the discrete space.
We select the autoregressive version because it was reported as the best alternative.  

\textbf{Experimental set-up. }We train all models using the RAdam optimizer~\cite{liu2019} and early stopping and keep the best model evaluated on the validation set. For the proteins dataset and for the large scale synthetic experiment $K=10$, in order to avoid overfitting, we monitor to ensure that the model is not reproducing more than $1\%$ of the training dataset in its generated samples. Performance metrics are averaged over 10 trials of $M= 10,000$ generated samples. A split of 70/20/10 is used for the protein datasets. 
The $p_{\theta}(\bX | \bZ^t, t)$ function is modeled using a non-autoregressive transformer similar to that used in~\cite{hoogeboom2021}. Following~\cite{ho2020}, we use sinusoidal position embedding to process the time step $t$ and concatenate it to $\bZ^t$ to form the input to the transformer.  The means $\bmu^*_1 ,\dots$ are computed using the procedure from~\cite{gamal1987} , which employs simulated annealing. We provide a complete description of architectures, the hyperparameters selection procedure in the supplementary. The source code is available at \url{https://github.com/networkslab/gmcd}.

\textbf{Results.} 
Experiments on the synthetic dataset highlight the modeling capability of GMCD. For every scale that we consider, $K=6,8,10$, GMCD outperforms at every distribution granularity: $\Omega, \mathcal{P}, \mathcal{P}^{od}$ (Table~\ref{tab:synth}). This is reflected in the covariance pattern metrics (Table~\ref{tab:synth_rho}).
The decomposition of $d_{TV}$ into the two regions $d_{TV+}$ and $d_{TVo}$ shows that most of the error for all baselines comes from  $d_{TV+}$, which is the error in estimating the probability mass of the valid sequences in $A_+$. 
This is to be expected as it is a harder task.
CDM is the closest competitor and its generated samples are almost all valid ($p(A_{+})$ is close to 100). Its deficiency is in assigning a probability mass ratio of approximately 2:1 to the two sets $A_{likely}$ and $A_{rare}$. This results in higher statistical distance metrics $d_{TV}$ and $Hel$. argmaxAR struggles to identify $A_+$ and requires additional training to reach a competitive result, but given more training time it can assign slightly better mass to the two sets, except for the larger scale experiment $K=10$. CNF is unable to distinguish between the likely and rare sets, which greatly impedes its performance for all metrics. 
As expected, as the problem grows harder, the fine-grained metrics $d_{TV}, Hel$ cannot be meaningfully estimated with this sample size.
For the protein dataset, GMCD is the best method overall and performs consistently for every pattern size (Table~\ref{tab:prot}).

\textbf{Ablation study and Time Analysis.} We report ablation studies to verify the relative contribution of two model components. We compare with a GMCD version with no sphere packing algorithm. The category distributions are randomly placed with no optimization (GMCD random). We also report GMCD trained with the initial sequence as in~\eqref{eqn:loss_almost_practice} (GMCD sharp). This eliminates data augmentation. As shown in the bottom of Table~\ref{tab:prot} the ablation experiment conducted on the PF00014 datasets confirms the relative importance of the components. We also include time and memory complexity of training and sampling of the models in the \textbf{abl.} section of Table~\ref{tab:timing}. GMCD requires the least time both for training and sampling because we can reduce the number of steps in the diffusion due to the more structured denoising procedure.
\section{Conclusion}
In conclusion, we introduced the GMCD model; a continuous diffusion-based model for nominal data. We introduced a novel novel fixed encoding procedure to map categorical data to the continuous space and gain representation flexibility. This also leads to a novel continuous denoising process that is cognizant of the categorical nature of the targeted distribution. The GMCD is fast to train, fast to sample from and generates representative samples of the ground truth distribution as demonstrated on synthetic and on a real world datasets.

\bibliographystyle{aaai}
\bibliography{ref}
\clearpage
\section{Diffusing Gaussian Mixtures for Generating Categorical Data \\ - Supplementary Material - }

\subsection{Derivation of the Gaussian Mixture component of the denoising process}

Here we provide the detailed derivation of the probability distribution of $\bZ^{t-1}$ conditioned on the end sequence $\bX$ and the current state $\bZ^t$ : $p(\bZ^{t-1} | \bZ^{t} , \bX)$ (Eqn. (4) from the main document). The solution is obtained from the following marginalization over $\bZ^0$: 
\begin{align}
     p(\bZ^{t-1} | \bZ^{t} , \bX) &=  \int p(\bZ^{t-1} | \bZ^{t} , \bZ^{0}) p( \bZ^{0}| \bX  )  d\bZ^0. 
\end{align}
 $p( \bZ^{0}| \bX  )$ is the encoder which we define as a m.v. Gaussian with means $[\bmu_{C_1}, \dots, \bmu_{C_K}]$ and standard deviation $ \sigma $ given by the sphere packing algorithm:
\begin{align}
 p( \bZ^{0}_{(s)}| x_{(s)}  ) &= \norm(\bZ^0_{(s)}  ; \bmu_{x_{(s)}}, \sigma^2 \mathbf{I}) ,\nonumber
\end{align}
and $p(\bZ^{t-1} | \bZ^{t} , \bZ^{0})$ is defined by the diffusion model: 
\begin{align}
p(\bZ^{t-1}_{(s)} | \bZ^{t}_{(s)} , \bZ^{0}_{(s)}) &= n(\bZ^{t-1}_{(s)} | \bZ^{t}_{(s)} , \bZ^{0}_{(s)}) \\&=  \norm(\bZ^{t-1}_{(s)}; \tilde{\bmu}_t, \tilde{\bbeta_t}), \nonumber \\ 
  \text{ where } \tilde{\bmu}_t &=  \frac{\sqrt{\bar{\alpha}_{t-1}}\beta_t}{1-\bar{\alpha}_t}\bZ_{(s)}^0+& \frac{\sqrt{\alpha_t}(1-\bar{\alpha}_{t-1})}{1-\bar{\alpha}_t}\bZ_{(s)}^t, \\ \tilde{\bbeta_t}&=\mathbf{I}\frac{1-\bar{\alpha}_{t-1}}{1-\bar{\alpha}_t}  \beta_t . 
  \end{align}
As a result, we are integrating over the product of Gaussians which has a closed-form solution:
\begin{align}
  p(\bZ^{t-1}_{(s)} &| \bZ^{t}_{(s)} , x_{(s)}) =  \int p(\bZ^{t-1}_{(s)} | \bZ^{t}_{(s)} , \bZ^{0}_{(s)}) p( \bZ^{0}_{(s)}| x_{(s)}  )  d\bZ^0.  \nonumber \\
  &= \int  \norm(\bZ_{(s)}^{t-1}; \tilde{\bmu}_t, \tilde{\bbeta_t}),  \norm(\bZ^0_{(s)}  ; \bmu_{x_{(s)}}, \sigma^2 \mathbf{I}) d\bZ^0  \nonumber,  \\
   \tilde{\bmu}_t& =  a \bZ^0+b \text{ hence } 
   \tilde{\bmu}_t \sim \norm(\tilde{\bmu}_t; a\bmu_{x_{(s)}} + b, (|a| \sigma)^2 \mathbf{I}) \end{align}
   so we can rewrite
  \begin{align} 
    p(\bZ^{t-1}_{(s)} &| \bZ^{t}_{(s)} , x_{(s)}) = \int  \norm(\bZ^{t-1}; \tilde{\bmu}_t, \tilde{\bbeta_t})
    \\ & \norm(\tilde{\bmu}_t; a\bmu_{x_{(s)}} + b, \mathbf{I}(|a| \sigma)^2 ) d\tilde{\bmu}_t  .
    \\
    \tilde{\bmu}_t =&  \frac{\sqrt{\bar{\alpha}_{t-1}}\beta_t}{1-\bar{\alpha}_t}\bZ_{(s)}^0+ \frac{\sqrt{\alpha_t}(1-\bar{\alpha}_{t-1})}{1-\bar{\alpha}_t}\bZ_{(s)}^t .
\end{align}
which has solution
\begin{align}
 p(\bZ^{t-1}_{(s)} | \bZ^{t}_{(s)} , x_{(s)}) &= \norm(\bZ^{t-1}_{(s)}; \bmu^{\bZ^t,t}_{x_{(s)}}, \bsigma^2_{t}) \nonumber \\
   \text{ where }\bmu^{\bZ^t,t}_{x_{(s)}}  &= \frac{\sqrt{\bar{\alpha}_{t-1}}\beta_t}{1-\bar{\alpha}_t} \bmu_{x_{(s)}} + \frac{\sqrt{\alpha_t}(1-\bar{\alpha}_{t-1})}{1-\bar{\alpha}_t}\bZ_{(s)}^t \nonumber \\
    \bsigma^2_{t} &= \mathbf{I}\frac{1-\bar{\alpha}_{t-1}}{1-\bar{\alpha}_t}  \beta_t+ \mathbf{I}(\frac{\sqrt{\bar{\alpha}_{t-1}}\beta_t}{1-\bar{\alpha}_t} \sigma)^2 . \nonumber 
\end{align}

\subsection{$\mathcal{L}_{t-1}$ approximation}
We provide the detailed derivation of the approximation for the individual loss terms $\mathcal{L}_{t-1}$.
In our architecture, $\mathcal{L}_{t-1}$ is the KL divergence between a Gaussian induced  by the diffusion process $n(\bZ^{t-1} | \bZ^{t}, \bZ^{0})$ and our denoising model $d_{\theta}(\bZ^{t-1} | \bZ^t)$; a product of Gaussian Mixtures with learnable mixture weights. 
\begin{align}
    n\left(\mathbf{Z}^{t-1} \mid \mathbf{Z}^{t}, \mathbf{Z}^{0}\right) &=  \norm(\bZ^{t-1}; \tilde{\bmu}_t, \tilde{\bbeta_t}) \\
    d_{\theta}(\bZ^{t-1} | \bZ^t) &=
  \prod_{s=1}^S \sum_{X'_{(s)}} p(\bZ^{t-1}_{(s)} | \bZ^{t}_{(s)} , X'_{(s)})  p_{\theta}(X'_{(s)}|\bZ^t, t)\, \nonumber
  \\&= \prod_{s=1}^S d_{(s),\theta}(\bZ_{(s)}^{t-1} | \bZ^t). 
\end{align}
The KL divergence between these two distributions is given by: 
\begin{align}
 \loss_{t-1}  &=KL\Big(n\left(\mathbf{Z}^{t-1} \mid \mathbf{Z}^{t}, \mathbf{Z}^{0}\right)||d_{\theta}(\bZ^{t-1} | \bZ^t)\Big) , \\
 &=  \sum^S_{s=1} KL\Big(n\left(\mathbf{Z}_{(s)}^{t-1} \mid \mathbf{Z}_{(s)}^{t}, \mathbf{Z}_{(s)}^{0}\right)|| d_{(s),\theta}(\bZ_{(s)}^{t-1} | \bZ^t) \Big),  \nonumber \\
 &= \sum^S_{s=1}  KL\Big(  \norm(\bZ^{t-1}_{(s)}; \tilde{\bmu}_t, \tilde{\bbeta_t})  || \\ &\sum_{X'_{(s)}}   p_{\theta}(X'_{(s)}|\bZ^t, t)\norm(\bZ^{t-1}_{(s)}; \bmu^{\bZ^t,t}_{X'_{(s)}}, \bsigma^2_{t})  \Big). 
 \end{align}

 \citeauthor{hershey2007} (2007) provide an approximation of the KL divergence between two Gaussian mixtures,  $f \triangleq \sum_a \pi_a \norm_a$ and $g \triangleq \sum_b w_b \norm_b $.
 The approximation relies on bounding the terms $\ex_f[\log g]$ and $\ex_f[\log f]$:
\begin{align}
\ex_f[\log g]  \geq& \ex_f \sum_b \phi \log \frac{w_b g_b}{\phi}
    \triangleq L_f(g, \phi) \text{ for some parameter } \phi.    \\
     \ex_f[\log f] \geq&  L_f(f, \zeta) \text{ for some parameter } \zeta. 
    \end{align}
   The parameters that maximize the bound,
  \begin{align}
\phi^*_{b,a} =& \frac{w_b e^{-KL(f_a||g_b)}}{\sum_b' w_{b'} e^{-KL(f_a||g_{b'})} } ,  \\
   \zeta^*_{a',a} =& \frac{\pi_{a'} e^{-KL(f_a||f_{a'})}}{\sum_{\tilde{a}}\pi_{\tilde{a}} e^{-KL(f_a||f_{\tilde{a}})} }  ,
    \end{align} 
 are then used to define the approximation:
\begin{align}
   KL(f||g)   \approx  KL_{var}(f||g) \triangleq& L_f(f, \zeta^*) - L_f(g, \phi^*). \nonumber \\
     =& \sum_a \pi_a \log \frac{\sum_{a'}\pi_{a'} e^{-KL(f_a||f_{a'})}}{\sum_b w_b e^{-KL(f_a||g_b)}} .
\end{align} 
In our case, since we have a single Gaussian as $f$ and not a mixture, the bound $  L_f(f, \zeta^*)$ becomes equal to the expectation $  L_f(f, \zeta^*) = \ex_f[\log f] $:
\begin{align}
    \zeta^*_{a',a} =& \frac{\pi_{a'} e^{-KL(f_a||f_{a'})}}{\sum_{\tilde{a}}\pi_{\tilde{a}} e^{-KL(f_a||f_{\tilde{a}})} } = 1 \nonumber  \\
    L_f(f, \zeta^*) =&  \ex_f \sum_a \zeta^* \log \frac{\pi_a f_a}{\zeta^*}  \nonumber \\ =& \ex_f \sum_a \log \pi_a f_a = \ex_f[\log f]   
\end{align}
Going back to $KL_{var}$ :
\begin{align}
    KL(f||g) &= \int \log \frac{f}{g} f \\
    &= \ex_f [\log f] - \ex_f [ \log g]  \nonumber  \\
    & \leq \ex_f [\log f] - L_f(g, \phi^*)  \nonumber \\
    KL(f||g) & \leq  KL_{var}(f||g)    \\
   &=  \sum_a \pi_a \log \frac{\sum_{a'}\pi_{a'} e^{-KL(f_a||f_{a'})}}{\sum_b w_b e^{-KL(f_a||g_b)}} \nonumber  \\
   &=  1 \log \frac{ 1 e^{-KL(f||f)}}{\sum_b w_b e^{-KL(f||g_b)}}  \nonumber \\
   & =  \log \frac{e^0}{\sum_b w_b e^{-KL(f||g_b)}}  \nonumber  \\
  \text{ hence we have }& KL(f||g) \leq  - \log \sum_b w_b e^{-KL(f||g_b)}. 
\end{align}
Applying this result to our loss, we obtain an expression that can be evaluated:
 \begin{align}
 \loss_{t-1}  &\leq  - \sum^S_{s=1}  \log  \\
 &\sum_{X_{(s)}'} p_{\theta}(X'_{(s)}|\bZ^t, t) \exp^{-KL(   \norm(\cdot ; \tilde{\bmu}_t, \tilde{\bbeta_t}) || \norm(\cdot; \bmu^{\bZ^t,t}_{X'_{(s)}}, \bsigma^2_{t} ) } .
 \end{align} 
 
\subsection{Algorithms}
Algorithm~\ref{algo:train} and Algorithm~\ref{algo:sampling} contain pseudocode for the training and sampling procedures, respectively.
For training, we introduced a ``diffused version'' $\tilde{\bx}^t$ of a sample sequence $\bx$. The closer we are to the beginning of the chain ($t=0$), the more mass in $p^{\bz^t, t }\big(\tilde{x}_{(s)} \big)$ will be concentrated to the category of $x_{(s)}$. The parameter  $\omega$ is a sharpening parameter that can increase or reduce that effect. As $\omega \to \infty$, $p^{\bz^t, t }\big(\tilde{X}_{(s)} = x_{(s)} \big) \to 1 $.
\begin{align}
    p^{\bz^t, t }\big(\tilde{\bX} \big)_s = \frac{ (w_{\bZ^{t,0}}^s(x_{(s)}) )^{\omega}}{\sum^K_{k=1} (w_{\bZ^{t,0}}^s(C_k) )^{\omega}}  \nonumber 
\end{align}
\begin{algorithm}[h]
\caption{Training}
\begin{algorithmic}[1]
\STATE {\bfseries Input:} Dataset $\data = \{\mathbf{x}^i\}^N_{i=1}, \bx_i \iid \bX$
\FOR{$\bx \in \data $}
\STATE Sample  $\bz^0 \sim q(\bZ|\bX = \bx)$.
\STATE Sample  $t \sim Uni(1, T)$ 
\STATE Sample $\bz^t \sim n(\bZ^t|\bZ^0 = \bz^0)$
\STATE Sample $\tilde{\bx}^t \sim p^{\bz^t, t }(\tilde{\bX}) $
\STATE Maximize $\log  p_{\theta}(\bX = \tilde{\bx}^t |\bZ^t=\bz^t, t)$(Eqn.[10] from the main document)
\ENDFOR
\RETURN $\theta$
\end{algorithmic}\label{algo:train}
\end{algorithm}

\begin{algorithm}[h]
\caption{Sampling }
\begin{algorithmic}[1]
\STATE {\bfseries Input:} Denoising $d_{\theta}(\bZ^{t-1}|\bZ^t, t)$, decoder $p(\bX|\bZ)$
\STATE {\bfseries Output:}  Sample of the trained model $ \bx \sim p_{\theta}(\bX)$
\STATE  Sample $\bz^T \sim \norm(\mathbf{0},\mathbf{1})$
\FOR{$t \in T, \dots 1 $}
\STATE Sample $\bx \sim p_{\theta}(\bX| \bZ^t=\bz^t, t)$ 
\STATE Sample $\bz^{t-1}_{(s)} \sim \norm(\bZ^{t-1}_{(s)}; \bmu^{\bZ^t,t}_{x_{(s)}}, \bsigma^2_{t}) $.
\ENDFOR
\RETURN $\bx \sim p(\bX|\bZ = \bz^0)$
\end{algorithmic}\label{algo:sampling}
\end{algorithm}

\subsection{Construction of the empirical pmf}\label{sec:empirical_poiss}
We describe how we obtain the empirical distribution from a set of samples generated by a generative model for evaluation purposes. 

Using the naive way of building an empirical distribution by using a shared fixed size sample set $m$ introduces unwanted dependencies between the estimates $\hat{p}_{\theta, x}$ of the  probability masses of different elements. 
To avoid this,  we obtain the empirical distribution $\hat{p}_{\theta}$ of a generative model $p_{\theta}$ through “Poissonization”. 

Instead of using the same number of samples $m$ to compute the frequency of appearance of some element $x$, for each $x \in \Omega$, we first sample the number of samples from a Poisson distribution $m' \sim Poi(m)$, and then construct the empirical pmf from a set of $m'$ samples of the generative model $\{\tilde{x}_i\}^{m'}_{i=1} ; \quad \tilde{x}_i \sim p_{\theta}$:
\begin{align}
  \hat{p}_{\theta, x} = \frac{1}{m'} \sum^{m'}_{i=1} \ind [\tilde{x}_i = x] .
\end{align}

\subsection{Sampling procedure for the patterns considered}\label{sec:rho}
Our evaluation of generative model includes computing the Pearson coefficient between lists of  pattern higher order covariations. Since we cannot evaluate the pattern higher order covariations for all possible patterns, we must make a selection which is described here.

For a given pattern length $p$, a total sequence length $S$ and a number of categories $K$, we sample likely patterns $\{pattern^p_1, \dots \}$ of all possible patterns following the procedure described in~\cite{mcGee2021}.
\begin{itemize}
    \item We first sample 1000 positions of size $p$ without replacement. We denote a position of size p by $pos^p = \{s_1, \dots, s_p\}$ where $ s_i \in \{1, \dots , S\}$ and $ s_i \neq s_j  \forall i\neq j$.We have $\{ pos^p_i\}^{1000}_{i=1} $.
    \item For each of these sampled $pos^p_i$, we find the top 20 most frequent patterns in the ground truth samples.  A pattern is a pair of positions associated to categories, $(pos^p, \{k_1, \dots, k_p\})$, hence we have $20$ patterns per $pos^p_i$ : $\{(pos_i^p, \{k_1, \dots, k_p\}_{i,j}), \}^{20}_{j=1}$.
    \item We combine all patterns of every 1000 positions to obtain a  list of 20,000 patterns $\{pattern^p_1, \dots pattern^p_{20k}\}$ that are used to compute the Pearson correlation $\rho^p$ between the pattern higher order covariations.
\end{itemize}

\subsection{Sphere packing}\label{sec:sphere_pack}
\begin{figure}[bht]
\centering
\begin{minipage}{.45\textwidth}
  \centering
    \includegraphics[scale=0.6]{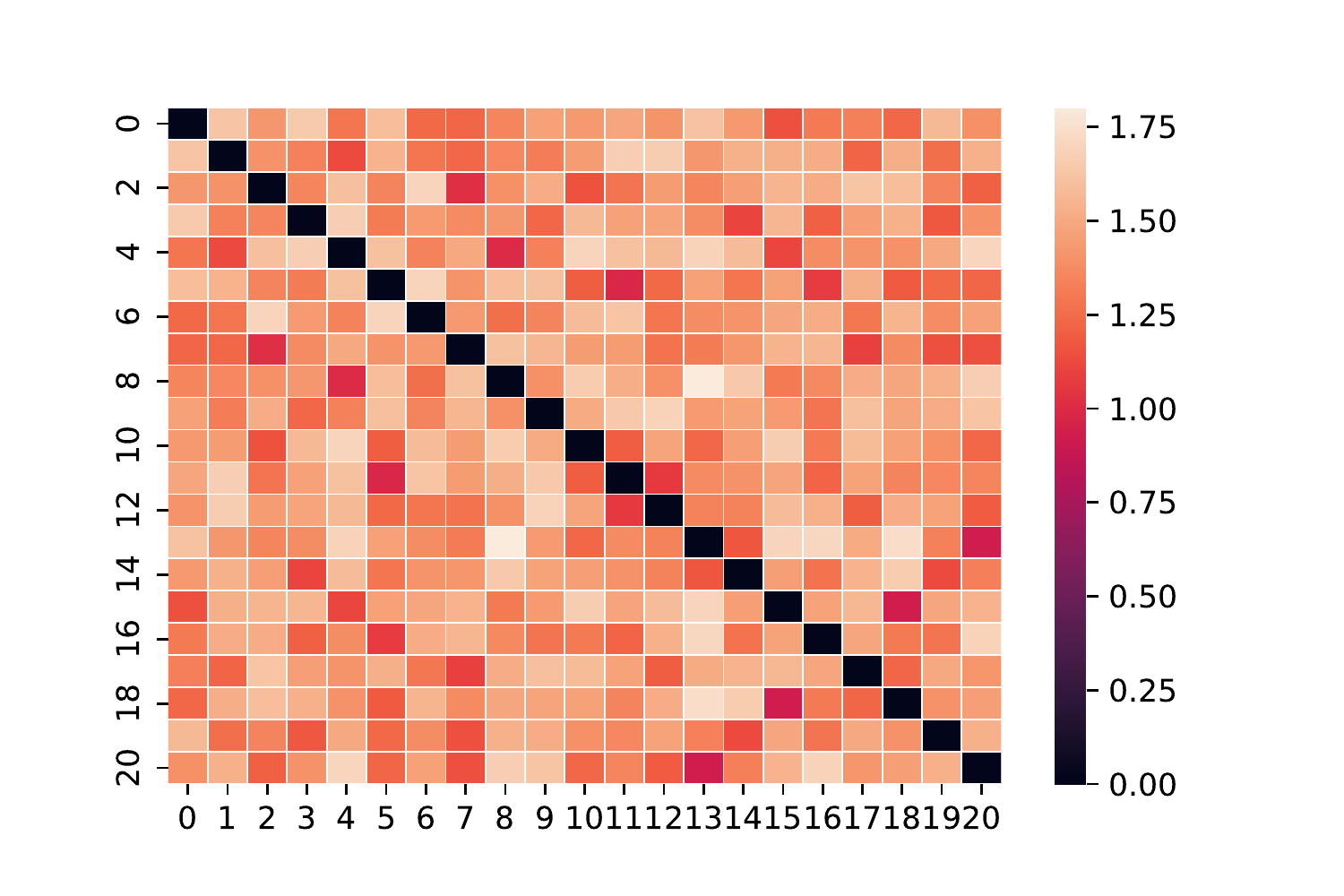}
\end{minipage}
\begin{minipage}{.45\textwidth}
  \centering
   \includegraphics[scale=0.6]{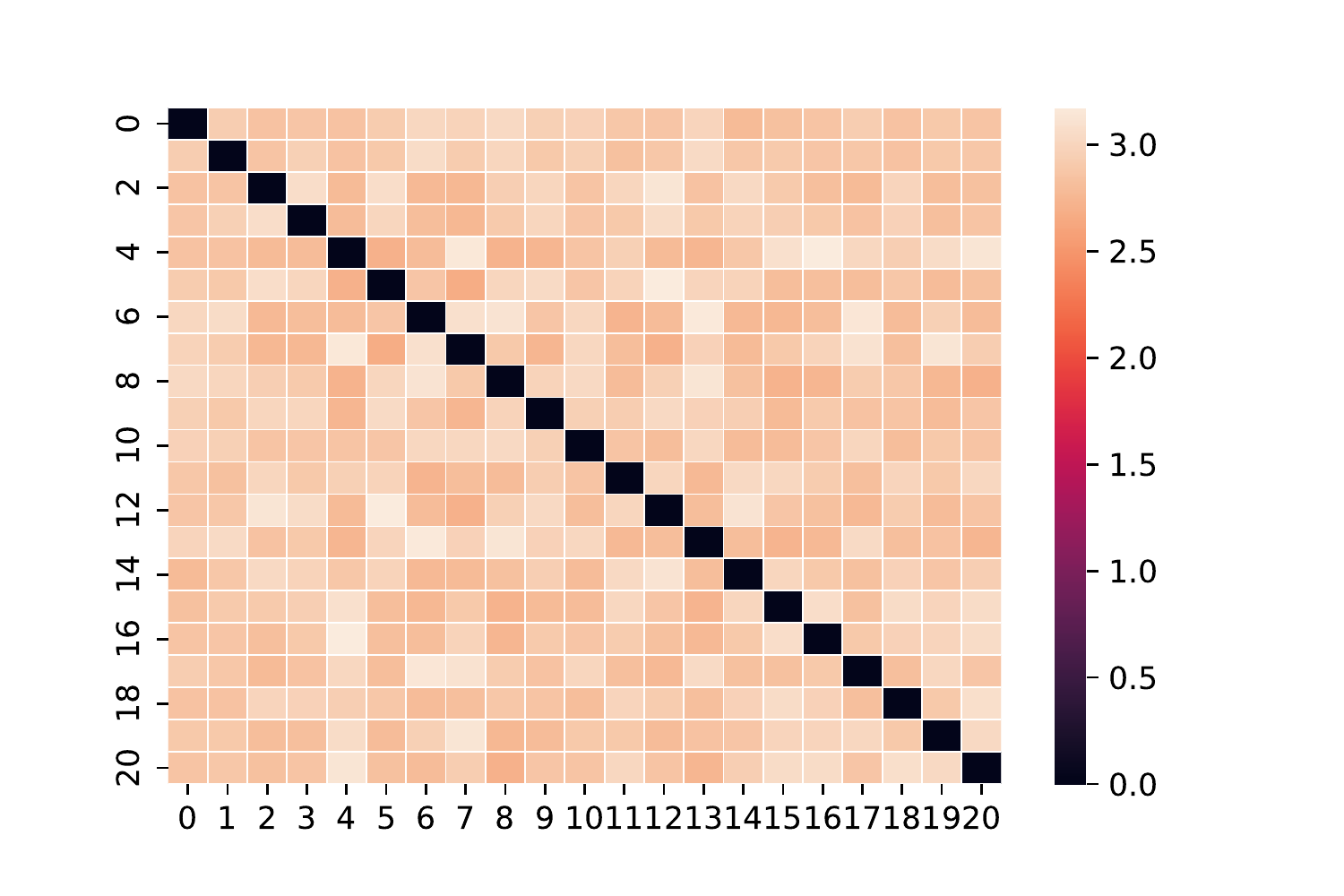}
\end{minipage}
\caption{$\ell$-2 distance between every pair of $\bmu^*_1 ,\dots, \bmu^*_K$ for $K=21$ of dimension $d=18$. The left figure shows the distances before the sphere packing algorithm, when the values are randomly initialized. The right figure shows the distances after the optimization. We can see that the algorithm successfully spreads out the means of every class representation  $\bmu^*_1 ,\dots, \bmu^*_K$, achieving a relatively even separation. Minimal bias is induced by having some classes placed closer than others. }\label{fig:sphere}
\end{figure} 
As explained in the methodology in the main paper, we use a sphere packing algorithm to determine the locations of the category representation in the continuous space. The goal is to obtain representations that are far away from each other.
\citeauthor{gamal1987} (1987) solve the problem through a simulated annealing based stochastic search to identify maximally separated points using an energy function that is the sum of the distances. 
The energy function $E(\bmu_1 ,\dots, \bmu_K)$ and perturbation function $ p( \bmu_1 ,\dots, \bmu_K) $ of the simulated annealing algorithm are given by:
\begin{align}
     E(\bmu_1 ,\dots, \bmu_K) &=  \sum_{i\neq k } ||\bmu_i - \bmu_j ||^2_2  \label{eqn:energy} ,\\
        p( \bmu_1 ,\dots, \bmu_K) &= \begin{cases}
     \bmu_i  &\text{ if }   i \neq k \\
      \bmu_i  + 0.1(\epsilon-0.5)  &\text{ if }   i = k
     \end{cases}   \label{eq:perturb}.\\
     \text{ with }& k \sim Uni[1,\dots, K], \epsilon \sim Uni(0,1]  
    \end{align}
    
The sphere packing algorithm and employed hyperparameters are described in Algorithm~\ref{algo:points}. 
The efficacy of the algorithm is shown in Figure~\ref{fig:sphere}

\begin{algorithm}[ht]
\caption{Sphere packing algorithm that solves for the means $\bmu^*_1 ,\dots, \bmu^*_K$}
\begin{algorithmic}[1]
\STATE {\bfseries Input:} Number of classes $K$, number of dimensions $d$
\STATE {\bfseries Output:}  means of the categories in the latent space $\bmu^*_1 ,\dots, \bmu^*_K \in \real^d$
\STATE Set $T=10, \alpha =0.9,MAX=100, \delta_{min}=0.001, MINSTEP=500$
\STATE $t=0, count=0$
\STATE Set initial values to random $\mathbf{U}_t = \bmu^t_1 ,\dots, \bmu^t_K \sim Uniform(\mathcal{S}^d(1))$
\STATE  Compute the energy $E_0 = E(\mathbf{\tilde{U}}_t)$ 
\WHILE{True }
  \STATE Add a small perturbation $\mathbf{\tilde{U}}_t = p( \mathbf{U}_t)$ (Eqn.~\eqref{eq:perturb})
  \STATE Compute the energy $E' = E(\mathbf{\tilde{U}}_t), \quad E = E(\mathbf{U}_t)$ (Eqn~\eqref{eqn:energy}) 
  \STATE $\delta_t = E' - E$
  \IF{ $\delta_t < 0$}
    \STATE Update  $\mathbf{U}_{t+1} = \mathbf{\tilde{U}}_t$
    \ELSE
    \STATE $coin \sim Bernouilli(p)$, with $p = e^{-\delta/T}$
    \IF{$coin =1$}
    \STATE Update  $\mathbf{U}_{t+1} = \mathbf{\tilde{U}}_t$
    \ENDIF
  \ENDIF
  \STATE $count= count+1$
  \IF{$count > max$}
  \STATE Reduce the temperature $T = T * \alpha$
  \STATE $count=0$
   \ENDIF
   
  \IF{$t > MINSTEP$ }
  \STATE Compute the average of the last 100 $\delta_t$ : $\bar{\delta} = \sum^{t}_{i=t-100} |\delta_i|$
  \IF{$\bar{\delta} < \delta_{min}$}
  \RETURN $\bmu^*_1 ,\dots, \bmu^*_K = \bmu_t^1, \dots, \bmu_t^K$
   \ENDIF
   \ENDIF
   \STATE $t= t+1$
 \ENDWHILE 

\end{algorithmic}\label{algo:points}
\end{algorithm}

\clearpage 
\subsubsection{Hyperparameter selection}\label{sec:hyper}
As stated in the paper, all experiments are trained with the RAdam optimizer, with a learning rate decay of $0.999975$, and parameters $\beta_1=0.9 $ and $\beta_2=0.999$. Tables~\ref{tab:fcdm_hyper},~\ref{tab:cdm_hyper},~\ref{tab:cnf_hyper}, and ~\ref{tab:argmax_hyper} report the architecture parameters for GMCD, CMD, CNF and argmaxAR are reported. For the CNF algorithm, we used the hyperparameters reported in~\cite{lippe2021categorical}.
For our proposed GMCD algorithm, to provide a fairer comparison, we employed the same hyperparameters for the transformer as the CDM~\cite{hoogeboom2021} where applicable. 
{\small \begin{table*}[h] \footnotesize
    \caption{GMCD hyperparameters.}
    \centering
    \begin{tabular}{lccccc} \toprule
       \textbf{Hyperparameters} &  \textbf{ $K=6$} & \textbf{ $K=8$} & \textbf{ $K=10$} &\textbf{PF00014} & \textbf{PF00076} \\ \midrule
         dim of $\bZ$ ($d$) &  $\{3,\dots, \underline{6}\}$& $\{3,\dots, \underline{8}\}$   &  \{ 3,\dots, \underline{9}, 10 \} &  \{ 3,\dots, \underline{15}, 21 \} & 15 \\  
          $\omega$ &  $\{1,5, \underline{\infty}\}$& $\{1,5, \underline{\infty}\}$  &  $\{\underline{1},5, \infty \}$ &  $\{\underline{1},5, \infty \}$ & 1 \\  \midrule 
        \textbf{transformer parameters} &  &    &  &&\\ \midrule
        hidden size & \{ 16, 32, \underline{64}\}  & \{ 16, \underline{32}, 64\}    & \{ 64, \underline{128}\}  & \{ 128, \underline{512}\}  & 512\\
        num. heads &8 &  8  & 8 &8 &8 \\
         depth &  2&  2  & 2 & 2 & 2 \\
         num. blocks & \{ 1, \underline{2}\} &  \{  \underline{1}, 2\} & 1 & \{  \underline{1}, 2\} & 1 \\
         local size & 64 & 64   & 64 &64 &64 \\
         local heads & 4& 4   & 4 &4&4\\\midrule
        dropout & 0.2 & 0.2 & 0.2& 0.2 & 0.2 \\
        T & \{\underline{10}, \dots, 50\} &  10  & 10  &  10 & 10\\
        batch size &  1024&  1024  & 1024 & 1024& 1024\\
        training iterations & 1k &  3k  &  2k & 10k & 30k \\
        learning rate &  \{7.5e-3, \underline{7.5e-4}, 7.5e-5\} &  7.5e-4 &  7.5e-4  & \{7.5e-3, \underline{7.5e-4}, 7.5e-5\} & 7.5e-4\\ \bottomrule
    \end{tabular}
    \vspace{1em}
    \label{tab:fcdm_hyper}
\end{table*}}

\begin{table*}[h]  \footnotesize
    \caption{CDM hyperparameters.}
    \centering
    \begin{tabular}{lccccc} \toprule
       \textbf{Hyperparameters} &  \textbf{ $K=6$} & \textbf{ $K=8$} & \textbf{ $K=10$} &\textbf{PF00014} & \textbf{PF00076} \\ \midrule
        \textbf{transformer parameters} &  &    &  &&\\ \midrule
        hidden size & \{ 16, 32, \underline{64}\}   & \{ 16, \underline{32}, 64\}    & \{ 64, \underline{128}\} & \{ 128, \underline{512}\}  & 512 \\
        num. heads &8 &  8  & 8 &8 &8 \\
         depth &  2&  2  & 2 &2 &2 \\
         local size & 64 &  64  & 64 &64&64\\
         local heads & 4 & 4   & 4 &4 &4 \\\midrule
        dropout & 0.2 & 0.2 & 0.2&0.2&0.2\\
        T & \{ 10, \underline{100}, 1000\}  & 100 & 100&100&100 \\
        batch size &  1024 &  1024  & 1024 &1024&1024\\
        training iterations & 1k & 3k &  2k &10k& 30k \\
        learning rate & \{7.5e-3, \underline{7.5e-4}, 7.5e-5\} &  7.5e-4&  7.5e-4  &\{7.5e-3, \underline{7.5e-4}, 7.5e-5\}& 7.5e-4\\ \bottomrule
    \end{tabular}
    \vspace{1em}
    \label{tab:cdm_hyper}
\end{table*}
\begin{table*}[h]  \footnotesize
    \caption{CNF hyperparameters. }
    \centering
    \begin{tabular}{lccccc} \toprule
          \textbf{Hyperparameters} &  \textbf{ $K=6$} & \textbf{ $K=8$} & \textbf{ $K=10$} &\textbf{PF00014} & \textbf{PF00076} \\ \midrule
        dim of $\bZ$ ($d$) &  \{2, \underline{5},6\} & \{2, \underline{3},8\}  & 5&\{2, \underline{5},6\} & 5 \\ \midrule
         \textbf{coupling parameters} &   & & & \\ \midrule
        num. layers & 2 &  2 & 2& 2&2  \\
         architecture & Transformer& Transformer& Transformer & Transformer & Transformer \\
        hidden size  & \{ 16, 32, \underline{64}\}  & \{ 16, \underline{32}, 64\}    & \{ 64, \underline{128}\}  & \{ 64,  \underline{128}, 512 \}  & 128\\
      num. mixtures &  \{2, \underline{4}, 8 \} & 4  & \{2, \underline{4}, 8 \} &\{2, \underline{4}, 8 \} & 4 \\
      mask ratio &  0.5  &  0.5  & 0.5  & 0.5& 0.5\\
      num. flows &  \{2, \underline{3}, 4 \}  &  3 &\{2, \underline{3}, 4 \} & \{2, 3, \underline{4} \} & 4  \\ \midrule
        batch size &  1024 & 1024  &1024 &1024 & 1024 \\
        training iterations & 4k & 5k  & 5k & 10k & 30k \\
        learning rate & \{7.5e-3, \underline{7.5e-4}, 7.5e-5\} &  7.5e-4 &7.5e-4 & \{7.5e-3, \underline{7.5e-4}, 7.5e-5\} & 7.5e-4\\ \bottomrule
    \end{tabular}
    \vspace{1em}
    \label{tab:cnf_hyper}
\end{table*}

\begin{table*}[h]  \footnotesize
    \caption{ArgmaxAR hyperparameters.}
    \centering
    \begin{tabular}{lccccc} \toprule
        \textbf{Hyperparameters} &  \textbf{ $K=6$} & \textbf{ $K=8$} & \textbf{ $K=10$} &\textbf{PF00014} & \textbf{PF00076} \\ \midrule
        encoder steps & \{\underline{2}, 3 , 4\} & 2 &\{\underline{2}, 3 , 4\} &\{2, 3 , \underline{4}\}&4 \\
        encoder bins &  \{2, \underline{4},5\}  & 4 & 4&5& 4\\
        context size &  \{16, 32, \underline{64}\}   & 64 & \{ \underline{64}, 128\} & \{128, \underline{256}\} & 256\\
        lstm layer & 1& 1 &1 &2& 1\\
        lstm size &  \{16, 32, \underline{64}\}  & 64 &  \{ 64, \underline{128}\} &\{128, \underline{256}\}& \{\underline{128}, 256\}\\
        context lstm layers & 1 & 1 &1  &1&1 \\
        context lstm size & \{16, \underline{32}, 64\}  & 32 &  \{16, \underline{32}, 64\} &  \{64 \underline{128}\}  & \{ \underline{64} , 128\} \\
        lstm dropout & 0.0 &  0.0  &0.0 & 0.0&0.0 \\
        batch size &  1024 & 1024 & 128&128& 128\\
        training iterations & 2k & 5k &6k &10k& 30k \\
        learning rate & \{7.5e-3, \underline{7.5e-4}, 7.5e-5\} &  7.5e-4&  7.5e-4  &\{7.5e-3, \underline{7.5e-4}, 7.5e-5\}& 7.5e-4\\ \bottomrule
    \end{tabular}
    \vspace{1em}
    \label{tab:argmax_hyper}
\end{table*}

\end{document}